\documentclass[acmsmall, screen, nonacm]{acmart}
\AtBeginDocument{%
  }

\setcopyright{none}
\renewcommand\footnotetextcopyrightpermission[1]{}

\usepackage{fontspec}

\newfontfamily\tamilfont[Path=./fonts/]{NotoSerifTamil-Regular.ttf}
\newfontfamily\japanesefont[Path=./fonts/]{NotoSerifJP-Regular.ttf}

\newcommand{\texttamil}[1]{{\tamilfont #1}}

\newcommand{\textjapanese}[1]{{\japanesefont #1}}

\newcommand{\paninian}{Pāṇinian}
\newcommand{\panini}{Pāṇini}
\newcommand{\astadhyayi}{Aṣṭādhyāyī}

\newenvironment{broadquote}
  {\list{}{
     \setlength{\leftmargin}{1em}
     \setlength{\rightmargin}{1em}
     \setlength{\parsep}{\parskip}
     \setlength{\topsep}{0.75em}
   }\item\relax\itshape}
  {\endlist}

\begin{document}

\title[A {\paninian} Foundation for Indic Language Processing]{A {\paninian} Foundation for Indic Language Processing}
\subtitle{One Metagrammar for a Billion Voices: Benchmarks and Architecture}

\author{Ritwik Banerjee}
\correspondingauthor
\email{rbanerjee@cs.stonybrook.edu}
\orcid{0000-0003-0336-0258}
\affiliation{%
  \institution{Department of Computer Science, Stony Brook University}
  \city{Stony Brook}
  \state{NY}
  \country{USA}
}
\additionalaffiliation{%
  \institution{AI Innovation Institute, Stony Brook University}
  \city{Stony Brook}
  \state{NY}
  \country{USA}
}

\author{Lav R. Varshney}
\email{lav.varshney@stonybrook.edu}
\orcid{0000-0003-2798-5308}
\affiliation{%
  \institution{AI Innovation Institute, Stony Brook University}
  \city{Stony Brook}
  \state{NY}
  \country{USA}
}
\additionalaffiliation{%
  \institution{Department of Electrical and Computer Engineering, Stony Brook University}
  \city{Stony Brook}
  \state{NY}
  \country{USA}
}


\begin{abstract}
More than a billion people communicate in Indic languages, yet the natural language processing infrastructure serving them remains fragmented and underdeveloped.
The cause is structural: the field organizes its tools and benchmarks around individual languages or small subsets of genealogical language families, building separate analyzers, parsers, and datasets for each language and starting over for the next.
This overlooks a deep regularity.
Through more than two millennia of convergence around Sanskrit, Indic languages came to share a morphosyntactic architecture formalized in {\panini}'s grammar, the {\astadhyayi}. This cuts across genealogical lines, uniting languages through a common framework.
We argue that this {\paninian} framework supplies a unifying computational architecture the field has lacked, and that benchmarks grounded explicitly in it would make Indic language systems more accurate, more data-efficient, and more transferable, effectively merging many apparently disparate and sparse Indic language resources into a single high-resource metalanguage bedrock.
We propose a four-part benchmark suite to render this shared architecture explicit, measurable, and ready to be leveraged for practical applications. Moreover, we underscore the question it raises for interpretability research: whether neural models trained on these languages come to represent {\panini}'s categories on their own.
\end{abstract}



\maketitle


\section{Introduction}
\label{sec:introduction}
Over a billion people across South Asia and beyond communicate daily in Indic languages --- Hindi, Bengali, Tamil, Telugu, Marathi, Santali, and dozens more --- yet the computational infrastructure serving these speakers/writers remains critically underdeveloped relative to its global significance.
As AI-powered tools for translation, accessibility, education, and information retrieval become central to modern life, the gap between what NLP can do for English or Chinese and what it can do for Indic languages carries real costs, both human and economic.
Closing that gap is an important engineering challenge facing the computing community today.

Progress, however, has been slow.
The primary reason is not a lack of data or talent, but a structural problem in how the field has organized itself.
The longstanding emphasis on genealogical divisions in the taxonomy of Indic languages has led to a fragmentation of computational approaches: separate morphological analyzers, parsers, and annotated datasets for each language family, or even each individual language.
The result is redundant engineering effort, acute resource constraints, and benchmark infrastructure that cannot transfer across languages.
The scale of this duplication is concrete: the Universal Dependencies \cite{demarneffe2021ud} collection alone, as of release 2.18 in May 2026 \cite{zeman2026ud}, maintains 24 separately annotated treebanks across 18 Indic languages\footnote{\href{https://universaldependencies.org}{universaldependencies.org}}, each built as an independent effort, and conversions between their annotation schemata are frequently lossy \cite{ravishankar2017universal, rai2025mapping}.
Every new language effectively requires building from scratch.

This is not a denial of the value or the tremendous engineering achievements of massively multilingual models \cite{nllbteam2024scaling}.
But their design objective is breadth, not depth: covering hundreds of languages optimizes translation fluency and coverage, not the morphosyntactic and semantic structure that distinguishes Indic languages, and no current benchmark would reveal the difference.
Such models do achieve cross-lingual transfer, but the units that carry it across Indic languages --- the shared and borrowed vocabulary, the parallel inflectional patterns, and the subwords that encode them --- are shared precisely because these languages are organized by a common {\paninian} architecture%
\footnote{{\panini} was a Sanskrit grammarian from around the fifth century BCE; his \textit{{\astadhyayi}} (``Eight Chapters'') specifies the language in some 4{,}000 ordered rules built from a formal metalanguage and explicit rule-precedence mechanisms. The parallel between this rule system and modern formal grammars was drawn early in the history of modern computing: writing in the Communications of the ACM, \citet{ingerman1967panini} proposed that Backus-Naur form be renamed ``{\panini}-Backus form''. The correspondence is at best cosmetic, however --- the {\paninian} rule formalism is in fact strictly more powerful than the context-free grammars described by BNF \cite{penn2012panini}.}%
, whether through inheritance, sustained contact, or independent convergence.
The transfer is therefore already running on {\paninian} rails; the models merely exploit them implicitly and incompletely, through incidental surface signals rather than the deeper structure reflected by those signals.
Empirical evidence bears the fingerprints of this shared architecture: cross-lingual transfer is stronger within Indic languages than between Indic and non-Indic languages even when script differences are controlled for \cite{bafna2023cross, nag2023transfer}, and morphological analyzers built on {\paninian} principles transfer across language families rather than merely within them \cite{goyal2016design, hellwig2016improving}.
Scale alone, however, quickly reaches a real ceiling.
A fixed-capacity model spread across many languages dilutes per-language quality, with the gains gravitating towards high-resource languages \cite{conneau2020unsupervised}, leaving most Indic languages underserved. Furthermore, generic subword tokenizers fragment morphologically rich Indic words rather than recovering their morphemes \cite{brahma2025morphtok}.

Making the shared substrate explicit converts this transfer into an interpretable and designed mechanism, so that the meaning of a lexeme can be transparently tracked as it is inflected and compounded across languages, rather than lost to subword fragmentation determined by ad hoc statistics.
Na\"{i}ve data sharing across Indic languages, with no {\paninian} grounding, already increases tagging accuracy \cite{pawar2023evaluating}, suggesting that there is ample room for empirical improvements through explicit grounding.
Since the transfer behavior of multilingual models already reveals that they implicitly learn a structural prior commonly held across languages, it is likely that giving Indic languages an interpretable grammar already known to be a common foundation will boost language comprehension across the board.

The fragmented, language-by-language approach and the opacity of multilingual models err in opposite directions around the same fact: the structural unity of Indic languages is real --- the former ignores it and pays in redundant effort, whereas the latter gains from it without ever recognizing its central role.
That unity is not a modeling artifact but the structural commons observed consistently by native multilingual users, across the boundaries of Indic languages and genealogies: remarkable parallels in morphological patterns such as verb conjugation, case marking, and agglutination; syntactic structures including postpositions and participial relatives; and shared frameworks for expressing agency, causation, aspect, and evidentiality.

Far from superficial lexical borrowings, these phenomena reflect deep architectural similarities engendered by over two millennia of linguistic convergence where Sanskrit has functioned as a formal intellectual ``metalanguage'' across South Asia and beyond.
This is not an isolated observation.
In other linguistic traditions, Latin shaped the prestige registers and grammatical self-conception of English \cite{blake1996history, lurie2023vernacular}.
In the case of Indic languages, however, the shared architecture runs considerably deeper: Tamil grammarians, working within their own tradition, arrived at structural conclusions that converge with {\paninian} categories --- a finding that is computationally significant precisely because it transcends genealogical boundaries.
Dialectal and register variations, too, operate along predictable continua rather than discrete breaks: dialects typically differ in phonological realization or lexical choice while preserving fundamental morphosyntactic architecture.
The widespread diglossia%
\footnote{Diglossia is the coexistence of two varieties of the same language throughout a community. Often, one form is the literary dialect (the ``high'' register; e.g., Katharevousa, which is heavily influenced by classical Greek, and used in official communications), and the other is a common dialect of everyday usage (the ``low'' register, e.g., Demotic Greek, which is the standard vernacular).\label{fn:diglossia}}
across Indic contexts --- Sanskrit-Prakrit, literary and colloquial Tamil, Hindi and Khariboli --- follows systematic alternations within shared structural constraints.

The computational treatment of Indic languages has long suffered from a denial of these regularities.
The result is a landscape in which the structural unity underlying these languages remains invisible to the tools meant to process them.
We argue that {\panini}'s grammatical framework, formalized in the {\astadhyayi} \cite{vasu1897}, provides a unifying computational architecture the field needs; and that building benchmarks explicitly grounded in this framework will unlock a new generation of more capable, resource-efficient, and transferable Indic language processing systems.

\section{The {\paninian} Framework as Unifying Architecture}
\label{sec:framework}
The unifying architecture we propose is not a modern invention. {\panini}'s {\astadhyayi}, composed around 500 BCE, is one of the most sophisticated formal grammatical systems in human history --- and for over two millennia, it functioned as the shared intellectual operating system of South Asian discourse, regardless of language.
Philosophy, law, science, and aesthetics across the subcontinent were all conducted within its formal categories.
Sanskrit was not merely one language among many; it was the metalanguage that supplied the ontological primitives, syntactic templates, and morphophonological regularities through which thought was organized and communicated across the entire region.

A computing professional might find the following analogy useful: think of Marathi and Tamil as having arisen from distinct kernels --- their genealogical origins differ --- but with their higher-level design patterns, discourse semantics, and formal structures built to the specifications of {\paninian} Sanskrit grammar.
Just as software components built to a common interface specification remain interoperable regardless of their underlying implementation, Indic languages built on this shared specification remain structurally compatible at the level that matters most for computation: morphology, syntax, and semantic organization.
This compatibility, rather than being a mere theory, is concretely manifested in texts such as \'Saiva Siddh\=anta (\texttamil{சைவ சித்தாந்தம்}) that exist simultaneously in Sanskrit and Tamil literary traditions, with the underlying semantic and argumentative architecture intact across both.

The analogy understates a property of {\panini}'s system that speaks with unusual directness to modern computational practice.
Any grammar formalism powerful enough to describe a natural language is, almost inevitably, powerful enough to \textit{overgenerate}, i.e., to admit strings that the language does not. \citet{penn2012panini} show that {\panini}'s formalism is no exception in its raw expressive power. What is exceptional, however, is the discipline imposed upon it.
Through rule precedence and a layer of meta-conventions governing how rules compete and apply, the {\astadhyayi} yields a \emph{single} derivation for every grammatical Sanskrit sentence --- disambiguation built into the architecture rather than bolted on afterward.

No generation system in the Chomskyan tradition has this property, and it is precisely this absence, \citet{penn2012panini} argue, that obliges modern NLP to reach for its heavy statistical and probabilistic machinery: the numerical methods are, in large part, a means of curbing the overgeneration these formalisms cannot restrain on their own.
Seen this way, the marvel of the {\astadhyayi} is not how many correct analyses it produces, but how many incorrect ones it entirely avoids.
An architecture grounded in this tradition would inherit determinacy as a core design principle. It does not obviate the statistical components of modern NLP, but provides a reason to expect that explicit {\paninian} structure carries information that modern systems recover only indirectly, and at a cost.

Demonstrated advantages of transfer learning within Indic languages \cite{bafna2023cross, nag2023transfer} and existing morphological analyzers based on {\paninian} principles \cite{goyal2016design, hellwig2016improving} support this picture directly.
Models trained on {\paninian} dependency labels show improved argument detection and semantic role labeling \cite{pal2019towards}.
Highly domain-specific tasks, such as translation of technical lexicon, show improvements in zero-shot setting when trained on Sanskrit tokens \cite{karthika2025levos}.

These are not marginal gains, but strong evidence that the shared architecture is computationally real and exploitable.

Furthermore, historical records support this picture independent of the computational rewards. Tamil grammarians did not merely borrow Sanskrit categories.
Rather, to a great extent, they discovered the same underlying categories independently within Tamil itself: Tolk\=appiyam, the oldest extant Tamil grammar text, describes \textit{sandhi} and ideas similar to \textit{vibhakti}; the 11th-century grammatical treatise V\=irac\=o\b{l}iyam incorporated k\=araka analysis and presented older traditional components of Tamil grammar through a comparative lens; and Ilakka\d{n}akkottu argued that Sanskrit grammatical features not mentioned in Tamil grammar nevertheless occur in the Tamil language, and therefore belong in its grammar \cite{annamalai2024sanskrit}.
These are records of recognition, not imposition \cite{acharya2013civilizations, pollock2000cosmopolitan}: two sophisticated grammatical traditions, arriving independently, and through interactions, at shared structural conclusions.
That convergence is precisely what enables computational leveraging of the {\paninian} framework across language families, not merely within a single genealogical tree (e.g., \citet{karthika2025multilingual} achieved improved tokenization by clustering Punjabi together with Dravidian languages); and it is what one would expect if the {\paninian} framework captures genuine computational primitives rather than being impositions of cultural prestige alone.

The relationship between genealogical and functional perspectives is worth exploring with precision, since the distinction matters for how we design computational systems.
Academic linguistics rightly focuses on genealogical descent to reconstruct historical development --- and it does not deny the structural commonalities we describe.
But genealogical taxonomy prioritizes inheritance, whereas computational transfer depends on functional and architectural similarity.
These are different questions, and they do not always have the same answer.
Recent work on cross-lingual transfer in programming languages makes this point sharply: genealogical relationships were not the most predictive features for transfer learning success; structural and corpus-specific features were far more reliable \cite{baltaji2025crosslingual}.
The same principle applies here.
The {\paninian} framework captures precisely the functional and architectural commonalities --- in morphology, syntax, and semantics --- that genealogical classification leaves implicit, and that computational systems need made explicit.

What this means practically is that the shared semantic infrastructure of {\paninian} grammar enables richer, more comprehensive knowledge representations for modern Indic languages.
The common syntactic templates enable transfer learning of foundational NLP processes --- semantic role labeling, tokenization, semantic composition, clause and phrase representation --- in ways that are both more efficient and more interpretable, because empirical methods can be grounded in an explicit, well-understood formal system.
Dialectal and register variation, rather than requiring separate handling, can be treated as parametric variation within this shared framework: dialects differ in phonological realization or lexical choice while preserving the underlying morphosyntactic architecture.
This is the key insight that the fragmented, language-by-language approach has been missing; and it points directly toward a new generation of unified, transferable benchmarks for Indic language technologies.

\section{Challenges in Computational Indic Language Processing}
\label{sec:challenges}
There are four distinct but interacting facets of Indic languages that challenge the development of effective Indic language technologies.

\subsection{Morphological complexity}
Indic languages are morphologically rich in ways that standard NLP pipelines --- designed primarily for English --- are poorly equipped to handle.
Words are not atomic units but structured compositions of roots and affixes encoding case, number, tense, mood, and aspect.
Two phenomena are particularly demanding.
\textit{Sandhi} --- the euphonic fusion of sounds at word boundaries --- means that segmenting a sentence into meaningful units is itself a non-trivial task requiring linguistic knowledge before any downstream processing can begin.
\textit{Samāsa}, the compounding of multiple semantic units into a single surface form, means that a single word may compress what English would express as an entire phrase.
These phenomena are pervasive in high-register communication and frequent even in everyday speech.
Handling them correctly is a prerequisite for almost every NLP task, yet most current tools treat them as edge cases rather than central architectural concerns.

\subsection{Diglossia, register variation, and code-mixing}
Across Indic language communities, speakers routinely navigate multiple registers --- literary and colloquial Tamil, Sanskrit-inflected Hindi, Khariboli --- and switch between them fluidly depending on context.
This diglossia is not noise, but a structural feature of how these languages function socially.
Code-mixing, particularly between an Indic language and English, adds a further layer of complexity that is especially pronounced on social media and in urban speech.
Current NLP models, trained predominantly on written standard varieties, struggle with all of these variations.
The deeper problem is that existing tools treat each register or mixed variety as a separate data distribution requiring its own handling, when in fact these variations follow systematic patterns within shared structural constraints.

The Hindi–Urdu pair makes this concrete.
In their everyday spoken forms the two are near-dialects of a single language, yet their formal registers diverge so sharply in vocabulary (Hindi drawing on Sanskrit, Urdu on Persian and Arabic) that the high varieties become mutually unintelligible \cite{king2006poisonous}.
What survives that divergence is the grammar: across the full register range the morphosyntactic scaffolding stays fixed, down to shared function morphemes such as the genitive \textit{-kī}.
The variation is large, but it is lexical and graphemic, layered over an invariant Indo-Aryan morphology and syntax.
That an entire prestige vocabulary can be swapped out while the structure holds is the sharpest demonstration that what these varieties share is architectural --- precisely the regularity that a register-by-register modeling approach leaves unexploited.

\subsection{The nominal semantics mismatch}
Perhaps the least appreciated challenge is a fundamental mismatch between the semantic assumptions embedded in standard NLP frameworks and the actual semantic organization of Indic languages.
Contemporary NLP, shaped by its development on Western languages, treats nouns as static, discrete entities whose meaning is grounded in extra-linguistic reference.
Indic languages, by contrast, encode a process-oriented ontology: nominal meaning is derivationally and conceptually rooted in verbal roots.
A noun, in Sanskrit --- and across Indic languages --- is best understood not as a static label but as denoting a role within an implicit event structure. The word \textit{dhātu}, for instance, means both ``root'' and ``metal'', but its {\paninian} definition translates as ``that which supports/holds'' --- an action-grounded description.
Standard benchmark tasks built on English-derived semantic assumptions systematically fail to evaluate this dimension of meaning, leaving a critical gap in how we measure language comprehension for Indic languages.

\subsection{Dataset scarcity and annotation incompatibility}
Even where datasets exist, they are fragmented and mutually incompatible.
The annotated resources that do exist were each developed for individual languages or for a single family, and even the attempts at a shared standard were never carried across the Indic family of languages.
The Indian Language Corpora Initiative (ILCI) parallel corpus \cite{jha2010tdil} is POS-annotated using the common BIS tagset for Indian languages \cite{sankaran2008common}; the most linguistically ambitious effort, the multi-layered Hindi/Urdu treebank built at IIIT-Hyderabad \cite{bhatt2009multi, palmer2009hindi} --- which covers what is essentially one Indo-Aryan language in two scripts \cite{king2006poisonous} --- pairs a {\paninian} \textit{kāraka} dependency layer with a PropBank\footnote{The Proposition Bank, or PropBank, is a shallow and broad foundational natural language processing resource created by \citet{palmer2005propbank}, that took a practical approach to semantic representation by adding a layer of predicate-argument information, or semantic role labels, to the syntactic structures of the Penn Treebank. Essentially, it provides sentence-level annotations for ``who did what to whom''.}-style predicate-argument layer adapted from English.
The common BIS tagset spans many languages, but only at the surface, while the treebank's deeper, {\paninian}-informed annotation never reaches beyond Hindi and Urdu.
No resource carries the deep {\paninian} annotations across the family boundary, and their schemata --- a POS tagset, a \textit{kāraka} dependency scheme, an English-derived semantic layer --- do not align with one another or with the universal scheme discussed next.

Universal Dependencies%
\footnote{Universal Dependencies --- available at \href{https://universaldependencies.org}{universaldependencies.org} --- is a collaborative, open project with more than 600 contributors who have built over 200 treebanks spanning more than 150 languages. It offers a unified scheme for annotating grammar (including lexical categories, morphological attributes, and syntactic relations) across diverse human languages.}
corpora exist for several Indic languages \cite{ravishankar2017universal}, but their universal scheme does not map onto the grammatical categories these languages actually use, making cross-lingual benchmark transfer difficult in practice. Sanskrit NLP, meanwhile, has concentrated on segmentation and lemmatization, with little work connecting these tasks to morpheme-level semantics: projects such as the Sanskrit Sembank \cite{hellwig2025sanskrit} assign WordNet synsets to words but do not encode the semantic roles --- \textit{kāraka} --- that are central to how meaning is organized in these languages. The cumulative result is that virtually no existing benchmark is designed around the structural unity of Indic languages; nearly all follow evaluation frameworks built for English, measuring what is easy to measure rather than what matters most.

\section{The State of the Art: Promising Signals, Fragmented Progress}
\label{sec:state-of-the-art}
The research community has not been idle.
Across morphology, syntax, and semantics, there are genuine advances in computational Indic language processing --- and a consistent pattern within them: when {\paninian} structure is explicitly exploited, performance improves.
The problem is that these advances remain isolated.
No one has connected them into a coherent, unified framework.
The field has the ingredients; what it lacks is the architecture.

Highly multilingual large language models (MLLMs), such as NLLB \cite{nllbteam2024scaling}, are tremendous engineering achievements that offer rigorous benchmarks within their scope.
But their scope is translation quality --- fluency, adequacy, toxicity identification, etc.; not morphological, morphosyntactic, or semantic role structures, cross-register performance, morpheme-level semantics, or dialectal robustness across intra-language variations.
These models do not resolve the underlying problem of fragmented corpora, and thus, the field's design limitations are continually inherited, not overcome.
The wide coverage of MLLMs is an engineering objective different from deep comprehension.
Without benchmarks designed to probe the underlying structural understanding of Indic languages, we cannot know how much these models have achieved, or how much more they may achieve when leveraging the universal substrate of {\paninian} metagrammar.

\subsection{Morphological analysis: strong foundations, limited reach}
The most mature body of work addresses morphological segmentation.
Word segmentation benchmarks built on \textit{sandhi} and \textit{samāsa} exist for Sanskrit \cite{krishna2017dataset}, and recent models such as ByT5-Sanskrit \cite{nehrdich2024one} achieve state-of-the-art results on segmentation and lemmatization.
The MorphTok benchmark \cite{brahma2025morphtok} grounds morphological analysis explicitly in {\paninian} grammar and has demonstrated downstream improvements in practice.
Earlier work established similar gains in machine translation \cite{banerjee2018meaningless} and named entity recognition \cite{pattnayak2025tokenization}.
The signal is clear and consistent: {\paninian} morphological grounding helps.
Yet these benchmarks are almost entirely confined to Sanskrit.
The extension to modern Indic languages, which share the same underlying morphological architecture, has not been done systematically.
A foundation exists, but it has not been built upon.

\subsection{Morphological tagging: cross-lingual gains left on the table}
Annotated corpora for morphological tagging exist for several modern Indic languages, but they trade breadth against depth.
The ILCI parallel corpus \cite{jha2010tdil} spans several Indo-Aryan as well as Dravidian languages --- Hindi, Bengali, Marathi, Tamil, Telugu, and others --- but stops at the shallow syntactic level of part-of-speech tagging.
The multi-layered treebank at IIIT-Hyderabad \cite{bhatt2009multi} annotates more, including gender, case, and number and a {\paninian} kāraka dependency layer, but reaches only the near-identical Hindi/Urdu language pair.
Pratyaya-Kosh \cite{singh2020benchmark} is itself grounded in {\paninian} \textit{pratyaya} analysis but covers only Sanskrit noun derivation.
These are valuable resources; the limitation they share is not that they ignore {\paninian} structure --- a few embrace it --- but that none provides unified {\paninian} morphological annotation across the Indic languages.
Breadth comes without depth, and depth without breadth, leaving the structural unity of Indic morphology fragmented across partial, mutually incompatible resources rather than exploited by a common scheme.
That even a fraction of this unity is exploitable is already demonstrated by \citet{pawar2023evaluating}, who trained a single multilingual model for morphosyntactic tagging, spanning both Indo-Aryan and Dravidian languages. That this was shown even with no explicit {\paninian} grounding at all is both encouraging and sobering: if sharing data alone yields empirical gains, the advantage from a unified {\paninian} annotation could be substantially larger.

\subsection{Syntactic parsing}
A {\paninian} schema exists, but remains isolated.
The AnnCorra project \cite{bharati2002anncorra} developed a dependency annotation schema explicitly grounded in {\paninian} \textit{kāraka} relations --- one of the most direct implementations of classical grammar in computational form.
The downstream benefits are real: incorporating \textit{kāraka} over universal dependencies has shown immediate improvements in Indic-language question-answering (QA) \cite{verma2023karaka}.
Yet AnnCorra remains largely an isolated effort.
Universal dependency corpora exist for some modern Indic languages \cite{ravishankar2017universal}, but their annotation schemata do not map to Sanskrit grammar, limiting cross-lingual transfer.
An initial effort to build {\paninian} universal dependencies was made by \citet{tandon2016conversion}, but has not been extended across multiple languages. The infrastructure for a unified multilingual {\paninian} parser is within reach --- the conceptual work has been done --- but the execution remains incomplete.

\subsection{Morpheme semantics: almost entirely uncharted}
The deepest gap is at the level of meaning. The Sanskrit Sembank \cite{hellwig2025sanskrit} assigns WordNet synsets to Sanskrit words --- a valuable resource --- but does not encode kāraka roles or root-level meanings. State-of-the-art segmentation models like ByT5-Sanskrit correctly identify lemmas and morphological tags, but make no connection to what those morphemes mean. There is virtually no work in the NLP literature that explicitly models the semantics of individual roots and affixes in Indic languages --- despite the fact that this morpheme-level semantic transparency is a defining feature of how these languages organize meaning. This is not a minor gap. It means that current systems can parse the surface form of an Indic sentence with reasonable accuracy while remaining blind to the semantic architecture that gives that sentence its meaning. No benchmark currently exists that would even reveal this blindness, let alone measure it.

\subsection{Cross-lingual transfer}
The signal is strong, but the infrastructure is lacking.
We underscored an important finding in recent literature, where \citet{bafna2023cross} and \citet{nag2023transfer} demonstrated that transfer across Indic languages yields markedly stronger performance than transfer from Indic to non-Indic languages.
This is precisely the pattern one would predict if the {\paninian} framework captures genuine computational structure shared across these languages.
That latent structure, however, is not what today's deployed systems actually exploit. The most recent benchmarks show transfer still riding on surface signal.
For instance, CorIL \cite{bhattacharjee2025coril}, a parallel corpus and translation evaluation spanning both Indo-Aryan and Dravidian languages, finds a performance hierarchy organized by script.
IndicTrans2, state-of-the-art for Brahmi-script languages, collapses on Perso-Arabic Sindhi with a near-zero BLEU score, whereas the massively multilingual NLLB \cite{nllbteam2024scaling} and BhashaVerse \cite{mujadia2025bhashaverse} score several times higher.
The larger models close this gap through sheer breadth of script coverage rather than any structural insight. The deciding variable is a surface property (in this case, the writing system), not the grammar shared by the languages.
Neither outcome reflects structural understanding; both track orthographic and lexical overlap.
The starkest case is the one that should be easiest: Urdu is, grammatically, nearly identical to Hindi \cite{king2006poisonous}, yet a model relying on Indic-script lexical sharing cannot bridge the script divide.
The shared morphosyntactic architecture, the very thing Hindi and Urdu hold in common, is precisely what these systems fail to utilize.
On the other hand, \citet{goyal2016design} and \citet{hellwig2016improving} demonstrated how morphological analysis based on {\paninian} principles can work across language families. A handful of multilingual question-answer datasets exist as well \cite{singh2025indic}.
But these remain isolated data points rather than a coordinated research program.
The empirical evidence for the {\paninian} hypothesis is accumulating --- yet the benchmark infrastructure needed for its systematic testing, along with deliberate extensions, does not exist.

There is a clear pattern across every dimension of Indic language processing, be it morphological analysis, syntactic parsing, semantic role labeling, or cross-lingual transfer.
The same story repeats: isolated advances, consistent positive signals when {\paninian} structure is exploited, and no unified framework connecting them.
What the field needs is not more isolated experiments but a coordinated benchmark suite that makes the shared {\paninian} architecture explicit, measurable, and exploitable across all Indic languages.

The benefits of such a unified approach are measurable. \citet{chang2024multilinguality} find that adding moderate amounts of multilingual data improves low-resource language modeling about as much as enlarging the low-resource corpus itself by up to a third, and that the gain is driven by the syntactic similarity of the added data, with vocabulary overlap mattering only marginally\footnote{This is precisely what separates beneficial multilinguality from the drawbacks of relentless scaling --- structurally aligned data adds signal, unrelated data dilutes it. This has been called the ``curse of multilinguality'' \cite{conneau2020unsupervised}.}.
The advantage, thus, lies in the structural core, not the lexical surface, consistent with evidence that large models encode grammatical organization along directions shared across languages \cite{brinkmann2025large} and align cross-lingually without depending on shared vocabulary \cite{conneau2020emerging}.
This is what makes the Indic case so favorable: Indic languages converge structurally across family lines with comparable morphological richness, through long contact, despite belonging to different genealogical families \cite{kakwani2020indicnlpsuite}, and cross-lingual transfer tracks exactly this morphological and structural similarity \cite{bankula2025cross}.
For Indian languages, exploiting this relatedness is already established practice: a substantial body of low-resource machine translation and transliteration leverages the orthographic and lexical substrate these languages share \cite{kunchukuttan2022machine}.
A common {\paninian} substrate would carry this deeper by making the shared morphosyntactic structure formal and explicit, so the additive advantage to low-resource languages, as demonstrated by \citet{chang2024multilinguality}, can operate across the entire group of Indic languages.
Thereby, separately scarce Indic language resources become, at the level that actually drives language comprehension and multilingual transfer, a single large hub.

\section{A Benchmark Suite for Indic Language Processing}
The gap analysis above reveals a consistent pattern: when {\paninian} structure is explicitly leveraged, performance improves.
But no coordinated benchmark infrastructure exists to make this leverage systematic and reproducible across languages.
We propose four thematic benchmark clusters that together would constitute a unified {\paninian} evaluation suite for Indic language processing.
Each cluster is (a) directly motivated by empirical results, and (b) designed to be multilingual from the ground up rather than extended \textit{post hoc} and \textit{ad hoc} to other languages.

\subsection{Morphological segmentation and tagging}
The first cluster addresses the most foundational level of Indic language processing: the decomposition of words into their {\paninian} roots and affixes, and the interpretation of what those morphemes mean. It provides the common skeleton to consolidate (a) multilingual morphological segmentation and tagging, (b) tasks on etymology and morpheme semantics, and (c) lexical semantics.

Existing Sanskrit segmentation benchmarks \cite{krishna2017dataset, nehrdich2024one} and the MorphTok benchmark \cite{brahma2025morphtok} provide an important starting point, but their coverage is almost entirely confined to Sanskrit.
The first task extends these benchmarks to modern Indic languages, requiring models to split sentences into {\paninian} roots (\textit{dhātu}) and affixes (\textit{pratyaya} and \textit{lakāra}) and label their grammatical and semantic functions.
The empirical case for this extension is already made: \citet{pawar2023evaluating} demonstrated a 7\% gain in morphosyntactic tagging accuracy from sharing data across Indo-Aryan and Dravidian languages --- without any explicit {\paninian} grounding.
Grounding the annotation explicitly in {\paninian} categories should yield further gains (cf. \citet{brahma2025morphtok}), since producing large ground-truth corpora is demonstrably feasible: finite-state systems for segmentation and morphological analysis \cite{huet2005functional, krishnan2019sanskrit} and rule-derivation engines that simulate the {\astadhyayi}'s rules \cite{mishra2009simulating} can generate analyses at scale, and their outputs can be validated and normalized into tagged gold corpora \cite{krishnan2023validation}.

The second task goes deeper: given an inflected word in a modern Indic language such as Marathi or Bengali, the task is to identify its Sanskrit \textit{dhātu} and interpret its core meaning.
This is morpheme-level semantic grounding, the testbed for the process-oriented nominal semantics that distinguishes Indic languages from the static noun-entity model assumed by standard NLP.
Cross-lingual word-sense disambiguation and semantic textual similarity tasks can be built on top of this etymological grounding by using resources like the Sanskrit Sembank \cite{hellwig2025sanskrit}.

\subsection{Syntactic parsing via {\paninian} dependencies}
This benchmark pursues a single, high-impact goal: a unified multilingual dependency parser grounded in \textit{kāraka} relations rather than universal dependencies.
The conceptual groundwork exists --- the AnnCorra project \cite{bharati2002anncorra} developed a \textit{kāraka}-based dependency schema, and \citet{tandon2016conversion} made an initial effort toward {\paninian} universal dependencies --- but neither has been extended systematically across multiple languages or language families.
The benchmark task here is both to build or convert dependency treebanks for multiple Indic languages using {\paninian} dependency labels, and to evaluate parsers trained on this unified schema against language-specific baselines.
A single robust multilingual parser, once available, can serve as infrastructure for virtually every downstream task across Indic languages.

\subsection{Semantic role labeling and inference across registers and languages}
This cluster addresses meaning at the sentence and discourse level.
It consolidates semantic role comprehension across Indic languages and language families, and tests model performance across dialectal continua and register variations across Indic languages.
The core benchmark annotates sentence pairs and QA examples across multiple Indic languages (Hindi, Marathi, Bengali, etc.) with \textit{kāraka} roles as semantic frames, analogous to PropBank but grounded in {\paninian} categories. Our proposal is to build directly on the \textit{kāraka}-grounded work by \citet{verma2023karaka}, extending it from QA to entailment and inference.

A distinct feature of these benchmarks is their explicit engagement with registers, as most Indic languages use multiple registers in both speech and text%
\footnote{The coexistence of formal and colloquial registers within a single language community (i.e., \textit{diglossia}, cf. footnote \ref{fn:diglossia}) is a defining feature of modern Indic language use. Bengali, for instance, distinguishes \textit{sādhu bhāṣā}, a highly Sanskritized written style, from \textit{cālitā bhāṣā}, the colloquial form used in most modern communication, except possibly in official documents. The gap between these registers is wider than, say, the difference between formal and informal English; it is closer to the difference between Latin and Italian. This pattern recurs across Indic languages: Tamil maintains a sharp divide between \textit{centamil} (literary) and \textit{koṭuntamil} (spoken); Hindi distinguishes a Sanskritized formal register from the colloquial Khaṛiboli. The phenomenon is not unique to India --- Arabic's \textit{fuṣḥā} versus \textit{'āmmiyya}, Japanese \textit{keigo} (\textjapanese{敬語}) versus plain speech, or Chinese classical/literary \textit{wényánwén} versus the vernacular \textit{báihuà}, are structural parallels --- but in the Indic context, register variation is especially significant for NLP because the formal register draws systematically on Sanskrit morphology and vocabulary, while the colloquial register reflects centuries of phonological and grammatical simplification.}%
.
Language models must capture the same underlying semantic content across these surface variations.
Benchmark tasks would require models to, for example, answer questions about a narrative given in colloquial Tamil that is derived from a Sanskrit original, or to identify entailment relations across literary and spoken register pairs.
These tasks directly test whether models have learned the underlying semantic architecture or merely memorized surface patterns.

\subsubsection*{Dialectal robustness and code-mixing generalization}
India exhibits an extremely rich set of local dialectal variations.
Tamil, for instance, exhibits significant variation across Chennai, Madurai, and Sri Lanka, and Bengali forms a comparable continuum --- from the southwestern districts of West Bengal through the northern belt around Cooch Behar to the eastern varieties of Tripura and Sylhet --- in phonology as well as morphology \cite{chatterji1926}.
These distinctions, much like the sharp register variations, share core morphosyntactic templates.
Key benchmark tasks ought to include cross-dialectal semantic role labeling, where training and test sets use different dialects or registers; register adaptation, requiring models to translate between literary and colloquial forms while preserving semantic content; and code-mixing generalization across sociolinguistic contexts ranging from formal written text to informal spoken interaction.
These tasks measure something that current benchmarks almost never measure: whether a model has learned structural invariants that generalize across surface variation, or whether it has simply memorized the patterns of a particular variety --- a distinction of enormous for real-world deployment across India's linguistic diversity.

\subsection{The extended Indic ``sprachbund'' and cross-lingual information disorder}
This final cluster pushes the unified framework in two directions simultaneously: outward to the broader Indic linguistic sphere, and toward an urgent social concern.
These benchmarks address what \citet{emeneau1956} identified as the \textbf{Indic sprachbund}\footnote{A ``sprachbund'' is an area of linguistic convergence, corresponding to a group of languages with similarities in syntax, morphological structure, cultural vocabulary, and sound systems \cite{thomason2000linguistic}.} and what one might think of, following the ``World Englishes'' framework of \citet{kachru1992other, kachru1992world}, as World Indic Languages.
This extension is striated, however, as one moves outward, and the benchmarks must respect that gradient.
At the first level are languages that remain Indo-Aryan and carry the full morphosyntactic substrate, however dispersed or divergent: diaspora varieties such as Caribbean Hindustani and Fiji Hindi, which preserve features lost in the modern standard; Romani, which retains Indo-Aryan morphosyntax despite centuries in Europe; and Sinhala and Dhivehi, the southernmost Indo-Aryan languages, which share the {\paninian} substrate along distinct developmental paths.
Here, the benchmark question is whether models trained on standard varieties generalize to these structurally Indic but divergent forms; and, conversely, whether the conservative among them, precisely because they preserve older features, can inform models about earlier stages of the shared {\paninian} architecture.
Then, there are languages connected not by structure but by civilizational influence.
Tibetan, Burmese, and Thai belong to different families altogether (Sino-Tibetan and Kra-Dai), yet they were written in Brahmi-derived scripts, inherited the Sanskrit phonological taxonomy that orders those scripts, and absorbed extensive Sanskrit vocabulary.
Herein lies a harder open question for the benchmark: how far can shared orthography, phonetic taxonomy\footnote{The {\paninian}/Śikṣā phonetic taxonomy is embedded in writing systems from Tibet to the farthest reaches of South-East Asia, including the Philippines.}, and lexicon --- in the absence of shared morphosyntactic substrate --- support transfer at all?
Together, these benchmarks establish how far the structural unity genuinely extends, and where it gives way to contact alone.

The applied direction addresses information disorder.
Misinformation in India spreads rapidly across linguistic boundaries, mutating as it moves between language communities --- a Hindi claim reappears in Punjabi with a subtle distortion, then in Bhojpuri or Bengali with another, each step compounding the original narrative while remaining semantically traceable to it.
Detecting and tracking this propagation requires models that understand the shared semantic substrate across languages, which is precisely what the {\paninian} framework provides.
Benchmark datasets for cross-lingual information disorder would require models to identify semantically equivalent claims expressed across multiple Indic languages and registers, track the variations introduced as content crosses linguistic boundaries, and flag the inconsistencies that signal deliberate manipulation.
For a computing community that has watched misinformation destabilize societies worldwide, building the infrastructure to detect it across one of the world's most linguistically complex regions is not just an academic exercise, but an engineering priority.

\section{Conclusion}
Nearly no existing benchmarks for Indic languages are designed around the structural foundation that actually unifies them.
Most follow evaluation frameworks built for English, measuring what is convenient rather than what is linguistically meaningful.
This article has argued that {\panini}'s grammatical framework, formalized over two millennia ago and active as an intellectual infrastructure across South and South-East Asia ever since, provides precisely the unifying computational architecture the field has been missing.
The architectural unity it captures is not a matter of genealogical inheritance alone: it extends even to Austroasiatic (e.g., Munda, whose speakers span India, Bangladesh, and Nepal) and Tibeto-Burman languages, where Sanskrit shaped not just the lexicon but the very phonological taxonomy that orders their scripts, and supplied the philosophical and technical vocabulary of their high registers.
This unity demonstrates that over two millennia of cultural-linguistic synthesis have created structural commonalities that run deeper than genealogical trees, and offer a framework for rapid development of practical AI-driven tools.

The practical payoff is substantial.
A multilingual parser trained jointly on Sanskrit, Hindi, and Marathi and outputting {\paninian} cases would be more accurate, more data-efficient, and more transferable than three separate parsers trained in isolation.
A multilingual question-answering benchmark where answers must be validated against \textit{kāraka} semantic roles would reveal capabilities and failure modes that remain invisible to current English-derived benchmarks.
The four benchmark clusters we have proposed would provide the evaluation infrastructure for a new generation of tokenizers, morphological analyzers, dependency parsers, and semantically grounded language models --- all built on a foundation that reflects how these languages actually work.

There is also a deeper scientific question at stake, one that should interest the computing community beyond its immediate engineering implications.
The {\paninian} framework's explicit formal structure --- \textit{kāraka} roles, \textit{dhātu} roots, morphological composition rules --- provides natural targets for mechanistic interpretability research.
We can directly probe whether neural models trained on Indic languages learn internal representations that align with {\paninian} categories, or whether they discover alternative organizational principles.
This raises a question analogous to the Platonic representation hypothesis in vision models \cite{huh2024platonic}:
\begin{broadquote}
Do neural language models trained on Indic languages spontaneously learn internal representations that correspond to {\paninian} categories, even without explicit supervision?
\end{broadquote}
If the answer is yes, it would mean that {\panini}'s framework does not merely offer a convenient analytical lens, but it captures genuine computational primitives of Indic linguistic structure, and it is the natural and canonical way in which morphosyntactic information is organized in (and across) these languages.
A framework conceived as a formal grammar would turn out to be a discovery about the deep structure of an entire sprachbund.

This prospect also reframes a methodological default the field rarely questions: if models already gravitate toward {\paninian} structure on their own, then leaning on statistical learning alone is less a first principle than a costly recovery mechanism from data.
There is a formal disambiguating structure that an explicit grammatical algebra supplies by design.
The most capable Indic language systems, then, are likely to come not from statistics alone, but from statistical models disciplined by such structure.

Building the benchmarks to answer this question is, in itself, a contribution to our understanding of how both artificial and human minds process language.
It is an invitation to the computing community to deeply engage with a rich and consequential linguistic tradition.


\bibliographystyle{ACM-Reference-Format}
\bibliography{references}

@book{acharya2013civilizations,
  title={{Civilizations in Embrace: The Spread of Ideas and the Transformation of Power : India and Southeast Asia in the Classical Age}},
  subtitle = {{The Spread of Ideas and the Transformation of Power : India and Southeast Asia in the Classical Age}},
  author={Acharya, Amitav},
  isbn={9789814379731},
  lccn={2012330954},
  series={Book collections on Project MUSE},
  year={2013},
  publisher={Institute of Southeast Asian Studies},
  address={Singapore}
}

@article{annamalai2024sanskrit,
  author = {E. Annamalai},
  title = {{The Sanskrit Paradigm of Tamil Grammar: Embrace and Resistance}},
  journal = {Bhasha},
  year = {2024},
  month = apr,
  volume = {3},
  number = {1},
  pages = {1--16},
  doi = {10.30687/bhasha/2785-5953/2024/01/002}
}

@inproceedings{bafna2023cross,
title = {{Cross-lingual Strategies for Low-resource Language Modeling: A Study on Five Indic Dialects}},
  author = "Bafna, Niyati and Espa{\~n}a-Bonet, Cristina and Van Genabith, Josef and Sagot, Beno{\^i}t and Bawden, Rachel",
  editor = "Servan, Christophe and Vilnat, Anne",
  booktitle = "Actes de CORIA-TALN 2023. Actes de la 30e Conf{\'e}rence sur le Traitement Automatique des Langues Naturelles (TALN), volume 1 : travaux de recherche originaux -- articles longs",
  month = "6",
  year = "2023",
  address = "Paris, France",
  publisher = "ATALA",
  url = "https://aclanthology.org/2023.jeptalnrecital-long.3",
  pages = "28--42"
}

@article{baltaji2025crosslingual,
  title={{Cross-lingual Transfer in Programming Languages: An Extensive Empirical Study}},
  author={Razan Baltaji and Saurabh Pujar and Martin Hirzel and Louis Mandel and Luca Buratti and Lav R. Varshney},
  journal={Transactions on Machine Learning Research},
  year={2025},
  volume = {2025},
  number = {June},
  numpages = {26},
  url={https://openreview.net/forum?id=1PRBHKgQVM}
}

@inproceedings{banerjee2018meaningless,
    title = {{Meaningless yet meaningful: Morphology grounded subword-level NMT}},
    author = "Banerjee, Tamali  and
      Bhattacharyya, Pushpak",
    editor = {Faruqui, Manaal  and
      Sch{\"u}tze, Hinrich  and
      Trancoso, Isabel  and
      Tsvetkov, Yulia  and
      Yaghoobzadeh, Yadollah},
    booktitle = "Proceedings of the Second Workshop on Subword/Character {LE}vel Models",
    month = jun,
    year = "2018",
    address = "New Orleans",
    publisher = "Association for Computational Linguistics",
    doi = "10.18653/v1/W18-1207",
    pages = "55--60"
}

@misc{bankula2025cross,
      title={Cross-Linguistic Transfer in Multilingual NLP: The Role of Language Families and Morphology}, 
      author={Ajitesh Bankula and Praney Bankula},
      year={2025},
      eprint={2505.13908},
      archivePrefix={arXiv}
}

@inproceedings{bharati2002anncorra,
  title = {{AnnCorra: Building Tree-banks in Indian Languages}},
    author = "Bharati, Akshar  and
      Sangal, Rajeev  and
      Chaitanya, Vineet  and
      Kulkarni, Amba  and
      Sharma, Dipti Misra  and
      Ramakrishnamacharyulu, K.V.",
    booktitle = "{COLING}-02: The 3rd Workshop on {A}sian Language Resources and International Standardization",
    year = "2002",
    url = "https://aclanthology.org/W02-1202",
    publisher = {Association for Computational Linguistics},
    address   = {Taipei, Taiwan},
    numpages = {8}
}

@inproceedings{bhatt2009multi,
  title = {{A Multi-Representational and Multi-Layered Treebank for Hindi/Urdu}},
  author = "Bhatt, Rajesh and Narasimhan, Bhuvana and Palmer, Martha and Rambow, Owen and Sharma, Dipti and Xia, Fei",
  editor = "Stede, Manfred and Huang, Chu-Ren and Ide, Nancy and Meyers, Adam",
  booktitle = "Proceedings of the Third Linguistic Annotation Workshop ({LAW} {III})",
  month = aug,
  year = "2009",
  address = "Suntec, Singapore",
  publisher = "Association for Computational Linguistics",
  url = "https://aclanthology.org/W09-3036",
  pages = "186--189"
}

@misc{bhattacharjee2025coril,
  title={{CorIL: Towards Enriching Indian Language to Indian Language Parallel Corpora and Machine Translation Systems}}, 
  author={Soham Bhattacharjee and Mukund K. Roy and Yathish Poojary and Bhargav Dave and Mihir Raj and Vandan Mujadia and Baban Gain and Pruthwik Mishra and Arafat Ahsan and Parameswari Krishnamurthy and Ashwath Rao and Gurpreet Singh Josan and Preeti Dubey and Aadil Amin Kak and Anna Rao Kulkarni and Narendra V. G. and Sunita Arora and Rakesh Balbantray and Prasenjit Majumdar and Karunesh K. Arora and Asif Ekbal and Dipti Mishra Sharma},
  year={2025},
  eprint={2509.19941},
  archivePrefix={arXiv}
}

@book{blake1996history,
  title={{A History of the English Language}},
  author={Blake, Norman},
  isbn={0-8147-1292-4},
  year={1996},
  publisher={New York University Press},
  address = {New York}
}

@preprint{brahma2025morphtok,
  title={{MorphTok: Morphologically Grounded Tokenization for Indian Languages}},
  author={Maharaj Brahma and N J Karthika and Atul Singh and Devaraj Adiga and Smruti Bhate and Ganesh Ramakrishnan and Rohit Saluja and Maunendra Sankar Desarkar},
  year={2025},
  eprint={2504.10335},
  archivePrefix={arXiv}
}

@inproceedings{brinkmann2025large,
    title = "Large Language Models Share Representations of Latent Grammatical Concepts Across Typologically Diverse Languages",
    author = "Brinkmann, Jannik  and
      Wendler, Chris  and
      Bartelt, Christian  and
      Mueller, Aaron",
    editor = "Chiruzzo, Luis  and
      Ritter, Alan  and
      Wang, Lu",
    booktitle = "Proceedings of the 2025 Conference of the Nations of the Americas Chapter of the Association for Computational Linguistics: Human Language Technologies (Volume 1: Long Papers)",
    month = apr,
    year = "2025",
    address = "Albuquerque, New Mexico",
    publisher = "Association for Computational Linguistics",
    doi = "10.18653/v1/2025.naacl-long.312",
    pages = "6131--6150",
}

@inproceedings{chang2024multilinguality,
    title = "When Is Multilinguality a Curse? Language Modeling for 250 High- and Low-Resource Languages",
    author = "Chang, Tyler A.  and
      Arnett, Catherine  and
      Tu, Zhuowen  and
      Bergen, Benjamin K.",
    editor = "Al-Onaizan, Yaser  and
      Bansal, Mohit  and
      Chen, Yun-Nung",
    booktitle = "Proceedings of the 2024 Conference on Empirical Methods in Natural Language Processing",
    month = nov,
    year = "2024",
    address = "Miami, Florida, USA",
    publisher = "Association for Computational Linguistics",
    doi = "10.18653/v1/2024.emnlp-main.236",
    pages = "4074--4096"
}

@book{chatterji1926,
  author = {Suniti Kumar Chatterji},
  title = {{The Origin and Development of the Bengali Language}},
  publisher = {Calcutta University Press},
  address   = {Calcutta, India},
  volumes   = {2},
  year = {1926}
}

@inproceedings{conneau2020unsupervised,
  title = "Unsupervised Cross-lingual Representation Learning at Scale",
  author = "Conneau, Alexis and Khandelwal, Kartikay and Goyal, Naman and Chaudhary, Vishrav and Wenzek, Guillaume and Guzm{\'a}n, Francisco and Grave, Edouard and Ott, Myle and Zettlemoyer, Luke and Stoyanov, Veselin",
  editor = "Jurafsky, Dan and Chai, Joyce and Schluter, Natalie and Tetreault, Joel",
  booktitle = "Proceedings of the 58th Annual Meeting of the Association for Computational Linguistics",
  month = jul,
  year = "2020",
  address = "Online",
  publisher = "Association for Computational Linguistics",
  url = "https://aclanthology.org/2020.acl-main.747/",
  doi = "10.18653/v1/2020.acl-main.747",
  pages = "8440--8451",
}

@inproceedings{conneau2020emerging,
    title = "Emerging Cross-lingual Structure in Pretrained Language Models",
    author = "Conneau, Alexis  and
      Wu, Shijie  and
      Li, Haoran  and
      Zettlemoyer, Luke  and
      Stoyanov, Veselin",
    editor = "Jurafsky, Dan  and
      Chai, Joyce  and
      Schluter, Natalie  and
      Tetreault, Joel",
    booktitle = "Proceedings of the 58th Annual Meeting of the Association for Computational Linguistics",
    month = jul,
    year = "2020",
    address = "Online",
    publisher = "Association for Computational Linguistics",
    doi = "10.18653/v1/2020.acl-main.536",
    pages = "6022--6034",
}

@article{demarneffe2021ud,
  author = {de Marneffe, Marie-Catherine and Manning, Christopher D. and Nivre, Joakim and Zeman, Daniel},
  title = {Universal Dependencies},
  journal = {Computational Linguistics},
  volume = {47},
  number = {2},
  pages = {255--308},
  year = {2021},
  month = jul,
  doi = {10.1162/coli_a_00402}
}

@article{emeneau1956,
  author = {Murray B. Emeneau},
  title = {{India as a Linguistic Area}},
  journal = {Language},
  volume = {32},
  number = {1},
  pages = {3--16},
  year = {1956},
  doi = {10.2307/410649},
  publisher = {Linguistic Society of America}
}

@article{goyal2016design,
  author = {Goyal, Pawan and Huet, Gerard},
  title = {{Design and analysis of a lean interface for Sanskrit corpus annotation}},
  journal = {Journal of Language Modelling},
  year = {2016},
  volume = {4},
  number = {2},
  pages = {145--182},
  doi = {10.15398/jlm.v4i2.108}
}

@inproceedings{hellwig2016improving,
  title = {{Improving the Morphological Analysis of Classical Sanskrit}},
  author = "Hellwig, Oliver",
  editor = "Wu, Dekai and Bhattacharyya, Pushpak",
  booktitle = "Proceedings of the 6th Workshop on South and Southeast {A}sian Natural Language Processing ({WSSANLP}2016)",
  month = dec,
  year = "2016",
  address = "Osaka, Japan",
  publisher = "The COLING 2016 Organizing Committee",
  url = "https://aclanthology.org/W16-3715/",
  pages = "142--151"
}

@article{hellwig2025sanskrit,
  author = {Oliver Hellwig and Erica Biagetti},
  title = {{The Sanskrit Sembank}},
  journal = {Language Resources and Evaluation},
  year = {2025},
  pages = {3635--3658},
  volume = {59},
  doi = {10.1007/s10579-025-09852-1}
}

@article{huet2005functional,
  author  = {Huet, G{\'e}rard},
  title   = {{A Functional Toolkit for Morphological and Phonological Processing, Application to a Sanskrit Tagger}},
  journal = {Journal of Functional Programming},
  volume  = {15},
  number  = {4},
  pages   = {573--614},
  year    = {2005},
  doi = {10.1017/S0956796804005416}
}

@inproceedings{huh2024platonic,
  title = 	 {{Position: The Platonic Representation Hypothesis}},
  author =       {Huh, Minyoung and Cheung, Brian and Wang, Tongzhou and Isola, Phillip},
  booktitle = 	 {Proceedings of the 41st International Conference on Machine Learning},
  pages = 	 {20617--20642},
  year = 	 {2024},
  editor = 	 {Salakhutdinov, Ruslan and Kolter, Zico and Heller, Katherine and Weller, Adrian and Oliver, Nuria and Scarlett, Jonathan and Berkenkamp, Felix},
  volume = 	 {235},
  series = 	 {Proceedings of Machine Learning Research},
  month = 	 {21--27 Jul},
  publisher =    {PMLR},
  url = 	 {https://proceedings.mlr.press/v235/huh24a.html},
  address = {Vienna, Austria}
}

@article{ingerman1967panini,
  author = {Ingerman, Peter Zilahy},
  title = {``Pānini-Backus Form'' suggested},
  year = {1967},
  publisher = {Association for Computing Machinery},
  address = {New York, NY, USA},
  volume = {10},
  number = {3},
  issn = {0001-0782},
  doi = {10.1145/363162.363165},
  journal = {Communications of the ACM},
  month = mar,
  pages = {137}
}

@inproceedings{jha2010tdil,
  title = {{The TDIL Program and the {I}ndian Langauge Corpora Intitiative (ILCI)}},
  author = {Jha, Girish Nath},
  editor = "Calzolari, Nicoletta and Choukri, Khalid and Maegaard, Bente and Mariani, Joseph and Odijk, Jan and Piperidis, Stelios and Rosner, Mike and Tapias, Daniel",
  booktitle = "Proceedings of the Seventh International Conference on Language Resources and Evaluation ({LREC}'10)",
  month = may,
  year = "2010",
  pages = {982--985},
  address = "Valletta, Malta",
  publisher = "European Language Resources Association (ELRA)",
  url = "https://aclanthology.org/L10-1602"
}

@book{kachru1992other,
  author = {Braj B. Kachru},
  title = {{The other tongue: English across cultures}},
  year = {1992},
  publisher = {University of Illinois Press},
  edition = {2},
  isbn = {978-0252062001},
  address = {Urbana, Illinois}
}

@article{kachru1992world,
  author = {Braj B. Kachru},
  title = {{World Englishes: approaches, issues and resources}},
  journal = {Language Teaching},
  volume = {25},
  number = {1},
  pages = {1--14},
  year = {1992},
  publisher = {Cambridge University Press},
  doi = {10.1017/S0261444800006583}
}

@inproceedings{kakwani2020indicnlpsuite,
    title = "{I}ndic{NLPS}uite: Monolingual Corpora, Evaluation Benchmarks and Pre-trained Multilingual Language Models for {I}ndian Languages",
    author = "Kakwani, Divyanshu  and
      Kunchukuttan, Anoop  and
      Golla, Satish  and
      N.C., Gokul  and
      Bhattacharyya, Avik  and
      Khapra, Mitesh M.  and
      Kumar, Pratyush",
    editor = "Cohn, Trevor  and
      He, Yulan  and
      Liu, Yang",
    booktitle = "Findings of the Association for Computational Linguistics: EMNLP 2020",
    month = nov,
    year = "2020",
    address = "Online",
    publisher = "Association for Computational Linguistics",
    doi = "10.18653/v1/2020.findings-emnlp.445",
    pages = "4948--4961"
}

@preprint{karthika2025multilingual,
  title={{Multilingual Tokenization through the Lens of Indian Languages: Challenges and Insights}}, 
  author={N. J. Karthika and Maharaj Brahma and Rohit Saluja and Ganesh Ramakrishnan and Maunendra Sankar Desarkar},
  year={2025},
  eprint={2506.17789},
  archivePrefix={arXiv},
  primaryClass={cs.CL}
}

@inproceedings{karthika2025levos,
  title = {{LEVOS: Leveraging Vocabulary Overlap with Sanskrit to Generate Technical Lexicons in Indian Languages}},
  author = "N J, Karthika and Bhatt, Krishnakant and Ramakrishnan, Ganesh and Jyothi, Preethi",
  editor = {Kochmar, Ekaterina and Alhafni, Bashar and Bexte, Marie and Burstein, Jill and Horbach, Andrea and Laarmann-Quante, Ronja and Tack, Ana{\"i}s and Yaneva, Victoria and Yuan, Zheng},
  booktitle = "Proceedings of the 20th Workshop on Innovative Use of NLP for Building Educational Applications (BEA 2025)",
  month = jul,
  year = "2025",
  address = "Vienna, Austria",
  publisher = "Association for Computational Linguistics",
  url = "https://aclanthology.org/2025.bea-1.20/",
  doi = "10.18653/v1/2025.bea-1.20",
  pages = "258--265",
  ISBN = "979-8-89176-270-1"
}

@article{king2006poisonous,
  title     = {{The Poisonous Potency of Script: Hindi and Urdu}},
  author    = {King, Robert D.},
  journal   = {International Journal of the Sociology of Language},
  volume    = {2001},
  number = {150},
  pages     = {43--59},
  year      = {2006},
  publisher = {Walter de Gruyter},
  doi = {10.1515/ijsl.2001.035}
}

@inproceedings{krishna2017dataset,
    title = {{A Dataset for Sanskrit Word Segmentation}},
    author = "Krishna, Amrith  and
      Satuluri, Pavan Kumar  and
      Goyal, Pawan",
    editor = "Alex, Beatrice  and
      Degaetano-Ortlieb, Stefania  and
      Feldman, Anna  and
      Kazantseva, Anna  and
      Reiter, Nils  and
      Szpakowicz, Stan",
    booktitle = "Proceedings of the Joint {SIGHUM} Workshop on Computational Linguistics for Cultural Heritage, Social Sciences, Humanities and Literature",
    month = aug,
    year = "2017",
    address = "Vancouver, Canada",
    publisher = "Association for Computational Linguistics",
    doi = "10.18653/v1/W17-2214",
    pages = "105--114",
}

@inproceedings{krishnan2019sanskrit,
    title = {{Sanskrit Segmentation revisited}},
    author = "Krishnan, Sriram  and
      Kulkarni, Amba",
    editor = "Sharma, Dipti Misra  and
      Bhattacharya, Pushpak",
    booktitle = "Proceedings of the 16th International Conference on Natural Language Processing",
    month = dec,
    year = "2019",
    address = "International Institute of Information Technology, Hyderabad, India",
    publisher = "NLP Association of India",
    url = "https://aclanthology.org/2019.icon-1.12/",
    pages = "105--114"
}

@inproceedings{krishnan2023validation,
    title = {{Validation and Normalization of DCS corpus and Development of the Sanskrit Heritage Engine{'}s Segmenter}},
    author = "Krishnan, Sriram  and
      Kulkarni, Amba  and
      Huet, G{\'e}rard",
    editor = "Kulkarni, Amba  and
      Hellwig, Oliver",
    booktitle = "Proceedings of the Computational {S}anskrit {\&} Digital Humanities: Selected papers presented at the 18th World {S}anskrit Conference",
    month = jan,
    year = "2023",
    address = "Canberra, Australia (Online mode)",
    publisher = "Association for Computational Linguistics",
    url = "https://aclanthology.org/2023.wsc-csdh.3/",
    pages = "38--58"
}

@book{kunchukuttan2022machine,
  author    = {Kunchukuttan, Anoop and Bhattacharyya, Pushpak},
  title     = {{Machine Translation and Transliteration involving Related and Low-Resource Languages}},
  publisher = {CRC Press},
  year      = {2022},
  isbn      = {978-0-367-56200-7},
  address = {Boca Raton, USA and Abingdon, UK}
}

@incollection{lurie2023vernacular,
  author = {David B. Lurie},
  title = {{The Vernacular in the World of Wen: Sheldon Pollock’s Model in East Asia?}},
  booktitle = {Cosmopolitan and Vernacular in the World of Wen \textjapanese{文}},
  year = "2023",
  month = may,
  publisher = "Brill",
  address = "Leiden, The Netherlands",
  isbn = "9789004529441",
  doi = "10.1163/9789004529441_003",
  pages = "49--68",
  editor = {King, Ross},
  type = {Chapter},
  volume = {5},
  series = {Language, Writing and Literary Culture in the Sinographic Cosmopolis}
}

@incollection{mishra2009simulating,
  author    = {Mishra, Anand},
  title     = {{Simulating the {\paninian} System of Sanskrit Grammar}},
  booktitle = {Sanskrit Computational Linguistics},
  editor    = {Huet, G{\'e}rard and Kulkarni, Amba and Scharf, Peter},
  series    = {Lecture Notes in Computer Science},
  volume    = {5402},
  pages     = {127--138},
  publisher = {Springer},
  year      = {2009},
  doi       = {10.1007/978-3-642-00155-0_4},
  address = {Berlin}
}

@preprint{mujadia2025bhashaverse,
  title={{BhashaVerse : Translation Ecosystem for Indian Subcontinent Languages}}, 
  author={Vandan Mujadia and Dipti Misra Sharma},
  year={2025},
  eprint={2412.04351},
  archivePrefix={arXiv}
}

@article{nag2023transfer,
  author = {Nag, Arijit and Samanta, Bidisha and Mukherjee, Animesh and Ganguly, Niloy and Chakrabarti, Soumen},
  title = {{Transfer Learning for Low-Resource Multilingual Relation Classification}},
  year = {2023},
  issue_date = {February 2023},
  publisher = {Association for Computing Machinery},
  address = {New York, NY, USA},
  volume = {22},
  number = {2},
  numpages = {24},
  pages = {1--24},
  doi = {10.1145/3554734},
  journal = {ACM Trans. Asian Low-Resour. Lang. Inf. Process.},
  month = mar,
  articleno = {50}
}

@inproceedings{nehrdich2024one,
    title = {{One Model is All You Need: ByT5-Sanskrit, a Unified Model for Sanskrit NLP Tasks}},
    author = "Nehrdich, Sebastian  and
      Hellwig, Oliver  and
      Keutzer, Kurt",
    editor = "Al-Onaizan, Yaser  and
      Bansal, Mohit  and
      Chen, Yun-Nung",
    booktitle = "Findings of the Association for Computational Linguistics: EMNLP 2024",
    month = nov,
    year = "2024",
    address = "Miami, Florida, USA",
    publisher = "Association for Computational Linguistics",
    doi = "10.18653/v1/2024.findings-emnlp.805",
    pages = "13742--13751"
}

@article{nllbteam2024scaling,
  author = {{NLLB Team}},
  title = {{Scaling neural machine translation to 200 languages}},
  journal = {Nature},
  volume = {630},
  pages = {841--846},
  year = {2024},
  doi = {10.1038/s41586-024-07335-x}
}

@inproceedings{pal2019towards,
  title = {{Towards Automated Semantic Role Labelling of Hindi-English Code-Mixed Tweets}},
  author = {Pal, Riya and Sharma, Dipti},
  editor = {Xu, Wei and Ritter, Alan and Baldwin, Tim and Rahimi, Afshin},
  booktitle = {Proceedings of the 5th Workshop on Noisy User-generated Text (W-NUT 2019)},
  month = nov,
  year = {2019},
  address = {Hong Kong, China},
  publisher = {Association for Computational Linguistics},
  doi = {10.18653/v1/D19-5538},
  pages = {291--296}
}

@article{palmer2005propbank,
  author    = {Martha Palmer and Paul Kingsbury and Daniel Gildea},
  title     = {{The Proposition Bank: An Annotated Corpus of Semantic Roles}},
  journal   = {Computational Linguistics},
  volume    = {31},
  number    = {1},
  pages     = {71--106},
  year      = {2005},
  doi       = {10.1162/0891201053630264}
}

@inproceedings{palmer2009hindi,
  author = {Martha Palmer and Rajesh Bhatt and Bhuvana Narasimhan and Owen Rambow and Dipti Misra Sharma and Fei Xia},
  title = {{Hindi Syntax: Annotating Dependency, Lexical Predicate-Argument Structure, and Phrase Structure}},
  booktitle = {Proceedings of the 7th International Conference on Natural Language Processing},
  series = {ICON},
  year = {2009},
  month = dec,
  pages = {259--268},
  address = {Hyderabad, India},
  publisher = {Macmillan Publishers}
}

@inproceedings{pattnayak2025tokenization,
  author={Pattnayak, Priyaranjan and Patel, Hitesh and Agarwal, Amit},
  booktitle={2025 IEEE International Conference on Electro Information Technology (eIT)}, 
  title={{Tokenization Matters: Improving Zero-Shot NER for Indic Languages}}, 
  year={2025},
  volume={},
  number={},
  pages={456--462},
  doi={10.1109/eIT64391.2025.11103625},
  publisher = {IEEE},
  address = {Valparaiso, Indiana, USA}
}

@inproceedings{pawar2023evaluating,
  title = {{Evaluating Cross Lingual Transfer for Morphological Analysis: a Case Study of Indian Languages}},
  author = "Pawar, Siddhesh and Bhattacharyya, Pushpak and Talukdar, Partha",
  editor = {Nicolai, Garrett and Chodroff, Eleanor and Mailhot, Frederic and {\c{C}}{\"o}ltekin, {\c{C}}a{\u{g}}r{\i}},
  booktitle = "Proceedings of the 20th SIGMORPHON workshop on Computational Research in Phonetics, Phonology, and Morphology",
  month = jul,
  year = "2023",
  address = "Toronto, Canada",
  publisher = "Association for Computational Linguistics",
  doi = "10.18653/v1/2023.sigmorphon-1.3",
  pages = "14--26"
}

@inproceedings{penn2012panini,
    title = {{On Pāṇini and the Generative Capacity of Contextualized Replacement Systems}},
    author = "Penn, Gerald  and
      Kiparsky, Paul",
    editor = "Kay, Martin  and
      Boitet, Christian",
    booktitle = "Proceedings of {COLING} 2012: Posters",
    month = dec,
    year = "2012",
    address = "Mumbai, India",
    publisher = "The COLING 2012 Organizing Committee",
    url = "https://aclanthology.org/C12-2092",
    pages = "943--950"
}

@article{pollock2000cosmopolitan,
  author = {Sheldon I. Pollock},
  title = {{Cosmopolitan and Vernacular in History}},
  journal = {Public Culture},
  year = {2000},
  volume = {12},
  number = {3},
  pages = {591--625},
  note = {Project MUSE},
  url = {https://muse.jhu.edu/article/26221}
}

@inproceedings{ravishankar2017universal,
  title = "A {U}niversal {D}ependencies Treebank for {M}arathi",
  author = "Ravishankar, Vinit",
  editor = "Haji{\v{c}}, Jan",
  booktitle = "Proceedings of the 16th International Workshop on Treebanks and Linguistic Theories",
  year = "2017",
  address = "Prague, Czech Republic",
  url = "https://aclanthology.org/W17-7623",
  pages = "190--200",
  publisher = {Association for Computational Linguistics}
}

@article{rai2025mapping,
  author = {Rai, Pooja and Das, Ayan and Chatterji, Sanjay},
  title = {Mapping of the Nepali Dependency Treebank to Universal Dependencies},
  year = {2025},
  issue_date = {November 2025},
  publisher = {Association for Computing Machinery},
  address = {New York, NY, USA},
  volume = {24},
  number = {11},
  pages = {1--22},
  numpages = {22},
  doi = {10.1145/3749643},
  journal = {ACM Trans. Asian Low-Resour. Lang. Inf. Process.},
  month = nov,
  articleno = {132}
}

@inproceedings{sankaran2008common,
  title = {{A Common Parts-of-Speech Tagset Framework for Indian Languages}},
  author = "Sankaran, Baskaran and Bali, Kalika and Choudhury, Monojit and Bhattacharya, Tanmoy and Bhattacharyya, Pushpak and Jha, Girish Nath and Rajendran, S. and Saravanan, K. and Sobha, L. and Subbarao, K.V.",
  editor = "Calzolari, Nicoletta and Choukri, Khalid and Maegaard, Bente and Mariani, Joseph and Odijk, Jan and Piperidis, Stelios and Tapias, Daniel",
  booktitle = "Proceedings of the Sixth International Conference on Language Resources and Evaluation ({LREC}'08)",
  month = may,
  year = "2008",
  pages = {1331--1337},
  address = "Marrakech, Morocco",
  publisher = "European Language Resources Association (ELRA)",
  url = "https://aclanthology.org/L08-1544",
}

@preprint{singh2020benchmark,
      title={{A Benchmark Corpus and Neural Approach for Sanskrit Derivative Nouns Analysis}}, 
      author={Arun Kumar Singh and Sushant Dave and Prathosh A. P. and Brejesh Lall and Shresth Mehta},
      year={2020},
      eprint={2010.12937},
      archivePrefix={arXiv}
}

@inproceedings{singh2025indic,
    title = {{INDIC QA BENCHMARK: A Multilingual Benchmark to Evaluate Question Answering capability of LLMs for Indic Languages}},
    author = "Singh, Abhishek Kumar  and
      Kumar, Vishwajeet  and
      Murthy, Rudra  and
      Sen, Jaydeep  and
      Mittal, Ashish  and
      Ramakrishnan, Ganesh",
    editor = "Chiruzzo, Luis  and
      Ritter, Alan  and
      Wang, Lu",
    booktitle = "Findings of the Association for Computational Linguistics: NAACL 2025",
    month = apr,
    year = "2025",
    address = "Albuquerque, New Mexico",
    publisher = "Association for Computational Linguistics",
    doi = "10.18653/v1/2025.findings-naacl.141",
    pages = "2607--2626"
}

@inproceedings{tandon2016conversion,
    title = {{Conversion from {\paninian} Karakas to Universal Dependencies for Hindi Dependency Treebank}},
    author = "Tandon, Juhi  and
      Chaudhary, Himani  and
      Bhat, Riyaz Ahmad  and
      Sharma, Dipti Misra",
    editor = "Friedrich, Annemarie  and
      Tomanek, Katrin",
    booktitle = "Proceedings of the 10th Linguistic Annotation Workshop held in conjunction with {ACL} 2016 ({LAW}-X 2016)",
    month = aug,
    year = "2016",
    address = "Berlin, Germany",
    publisher = "Association for Computational Linguistics",
    doi = "10.18653/v1/W16-1716",
    pages = "141--150"
}

@incollection{thomason2000linguistic,
  author       = {Thomason, Sarah Grey},
  title        = {{Linguistic Areas and Language History}},
  booktitle    = {{Languages in Contact}},
  editor       = {Gilbers, D. G. and Nerbonne, J. and Schaeken, J.},
  series       = {Studies in Slavic and General Linguistics},
  volume       = {28},
  pages        = {311--327},
  year         = {2000},
  publisher    = {Brill},
  address      = {Leiden, The Netherlands},
  doi = {10.1163/9789004488472_030}
}

@book{vasu1897,
  author = {Vasu, Srisa Chandra},
  title = {{The Ashtādhyāyī of {\panini}}},
  publisher = {Sindhu Charan Bose},
  address = {Benares},
  year = {1897}
}

@inproceedings{verma2023karaka,
    title = {{K\={a}raka-Based Answer Retrieval for Question Answering in Indic Languages}},
    author = "Verma, Devika  and
      Joshi, Ramprasad S.  and
      Shivani, Aiman A.  and
      Gupta, Rohan D.",
    editor = "Mitkov, Ruslan  and
      Angelova, Galia",
    booktitle = "Proceedings of the 14th International Conference on Recent Advances in Natural Language Processing",
    month = sep,
    year = "2023",
    address = "Varna, Bulgaria",
    publisher = "INCOMA Ltd., Shoumen, Bulgaria",
    url = "https://aclanthology.org/2023.ranlp-1.129",
    pages = "1216--1224"
}

@artifactdataset{zeman2026ud,
 title = {{Universal Dependencies 2.18}},
 author = {Zeman, Daniel and Nivre, Joakim and Abid, Rimsha and Abrams, Mitchell and others},
 url = {http://hdl.handle.net/11234/1-6149},
 note = {Institute of Formal and Applied Linguistics ({{\'U}FAL}), {LINDAT}/{CLARIAH}-{CZ} Digital Library},
 copyright = {Licence Universal Dependencies v2.18},
 year = {2026}
}

\end{document}